\documentclass[journal]{IEEEtran}
\usepackage[utf8]{inputenc}
\usepackage[T1]{fontenc}
\usepackage{newtxtext}
\usepackage{graphicx,array,amsmath}
\usepackage[hidelinks]{hyperref}

\begin{document}
\title{How Much of SAR Vehicle Recognition Is the Vehicle? Background Shortcuts, Angle Gaps and Partial Libraries on MSTAR and ATRNet-STAR}
\author{Mousa~Shaya%
\thanks{M. Shaya is with Shaya Software and Systems LLC and with Arizona State University, Tempe, AZ, USA (e-mail: mshaya01.ml@gmail.com).}}
\maketitle
\begin{abstract}
Deep networks score near 99\% on the MSTAR vehicle benchmark, and the newer ATRNet-STAR benchmark was built to be harder. We ask how much of these scores comes from the vehicle itself. With a modern network (ConvNeXt-Small, 3 seeds) we first reproduce the known MSTAR result: with the central 96\ensuremath{\times}96 pixels zeroed out, accuracy is still 73.3\% (chance 10\%). The same shortcut is present in ATRNet-STAR, which was collected across several scenes: with the vehicle removed, accuracy is 63.3\% on the 40-class SOC split (chance 2.5\%). Under the official scene-change split, the background stops helping and starts hurting. Accuracy is 22.8\% on full chips and 30.8\% when only the vehicle is shown. A SAR foundation model, SARATR-X, leans on the background more than ConvNeXt does (+4 to +6 points with the vehicle removed, in every seed) and does not solve the scene change. We then hold the amount of training data fixed and vary the depression angles used for training. Accuracy is set mainly by the gap to the nearest training angle: 57\% at 15\ensuremath{^\circ}, 39\% at 30\ensuremath{^\circ} and 26\% at 45\ensuremath{^\circ}. Adding more training angles does not help. Finally, when most classes have training images at the test angle and a few do not, the missing classes do worse than when no class has them: 9.0\% instead of 22.4\% at a 45\ensuremath{^\circ} gap. Most of their images are assigned to known classes. We recommend reporting background-only and target-only controls, and the angle gap, next to any accuracy figure. Code is public.
\end{abstract}

\section{Introduction}

Synthetic aperture radar (SAR) images the ground in any weather, day or night. This makes automatic target recognition (ATR) from SAR useful in practice, and it has been studied for decades. For more than 25 years the main yardstick has been MSTAR, a public set of X-band images of ten military vehicles. Under its standard operating condition (SOC), deep networks now score about 99\%.

A high score does not prove that a network recognizes the vehicle. Networks learn whatever separates the classes in the training data, and if something other than the object does the job, they use it \cite{geirhos2020}. In MSTAR the clutter around each vehicle is such a signal. This has been known for a long time. Schumacher and colleagues showed in 2004 and 2005 that several ATR methods score high by exploiting the shared clutter in MSTAR training and test images \cite{schumacher2004,schumacher2005}. Later work showed that CNN saliency falls on the clutter for most classes \cite{heiligers2018}, that a network trained on clutter alone beats chance \cite{zhou2018,belloni2021}, and measured how much each region contributes \cite{li2023}.

What is less clear is whether the problem goes away with newer data and newer models. ATRNet-STAR \cite{liu2026} is far larger than MSTAR: 40 civilian vehicle types, four depression angles, two bands and five scenes. Its authors contrast it with MSTAR's single grass field and its ``pseudo-correlation between target and background''. Its official scene-change split, which trains on simple scenes and tests on cluttered ones, gives very low accuracy, and the authors suggest that background interference is the cause. SAR foundation models such as SARATR-X \cite{li2025}, pretrained on large collections of SAR images, are also expected to learn better SAR features than networks pretrained on photographs.

We test both expectations with simple controls: show the network only the background, or only the vehicle, and see what accuracy remains. We then turn to a second practical issue, the depression angle. Real systems know the collection geometry and compare against reference libraries, and those libraries always have gaps. We measure how much accuracy an angle gap costs when the amount of training data is held fixed, and what happens when the library is complete for most vehicles but missing for a few.

This paper makes four contributions. First, we run background-only and target-only controls on ATRNet-STAR, using its bounding boxes. The background shortcut is present in this benchmark too (63.3\% with the vehicle removed), and under the official scene change, removing the background raises accuracy by 8 points. This directly measures what the dataset authors suspected. A control that sees only the mask shape rules out a masking artifact, and MSTAR is included as a reproduction of the known effect with a modern network. Second, we repeat the controls with a SAR foundation model. SARATR-X relies on the background more than an ImageNet-pretrained ConvNeXt, and it does not solve the scene change; as far as we know, this is the first such test of a SAR foundation model. Third, we study depression angle at a fixed data budget, with 14 sets of training angles, every held-out angle and 3 seeds. Accuracy depends on the gap to the nearest training angle, not on the number of training angles. Fourth, we describe a partial-library effect: when other classes have training images at the test angle, a class without them gets worse, and its images are mostly assigned to classes in the library.

All code is public, and every result is a mean over three seeds.

\section{Related Work}

\textbf{MSTAR and its protocols.} MSTAR was released by AFRL and DARPA in the late 1990s. The standard protocols, SOC and several extended operating conditions (EOC), were described by Ross et al.~\cite{ross1998}. Under SOC, networks train at 17\ensuremath{^\circ} depression and test at 15\ensuremath{^\circ}, and accuracy has been near 99\% for several years.

\textbf{Background bias in MSTAR.} This is well established, and we do not claim it as new. Schumacher and Rosenbach \cite{schumacher2004} and Schumacher and Schiller \cite{schumacher2005} showed that several ATR methods reach high scores by using the identical clutter in training and test images, and warned against extrapolating MSTAR results to real conditions. Heiligers and Huizing \cite{heiligers2018} used Grad-CAM on a CNN with 97.6\% accuracy and found that it relied on clutter for 6 of the 10 classes. Using segmentation masks, they reported 97.2\% for target-only input and 77.0\% for shadow-only input. Zhou et al.~\cite{zhou2018} trained a network on clutter alone and reached over 40\%. Belloni et al.~\cite{belloni2021} found that removing the clutter dropped a trained classifier from 98\% to 33\%, almost as much as masking the target (27\%). Li et al.~\cite{li2023} measured the contribution of target, shadow and clutter with Shapley values and traced the bias to similar signal-to-clutter ratios across classes. Several methods try to reduce the bias: training on segmented targets \cite{heiligers2018}, swapping in other clutter as augmentation \cite{geng2023}, contrastive clutter invariance \cite{peng2023} and causal regularization \cite{debiased2023}. Geng et al.~report that accuracy falls to 50--60\% when test clutter is swapped. All of these studies use MSTAR, most with smaller CNNs and single runs. We reproduce the effect with a modern network and three seeds, and then test the newer benchmark and a SAR foundation model.

\textbf{Newer benchmarks and foundation models.} ATRNet-STAR \cite{liu2026} is the largest public SAR vehicle set we know of, with official SOC and EOC splits (scene, depression, azimuth, band, polarization). SARATR-X \cite{li2025} is a HiViT encoder pretrained with masked image modelling on about 180,000 SAR images. A related line uses joint-embedding prediction for SAR ATR \cite{li2024}. Vision--language models for SAR have also appeared, including SARCLIP \cite{jiang2025}. In early tests, frozen CLIP and SARCLIP features fell well below fine-tuned networks on MSTAR, so we use SARATR-X as our SAR-pretrained model.

\textbf{Shortcut learning.} Geirhos et al.~\cite{geirhos2020} describe the general pattern: networks take the easiest available cue, which often is not the one intended. Background controls like ours are a standard way to detect it.

\section{Data and Methods}

\subsection{Datasets}

\textbf{MSTAR.} We use the public release with 10 classes (2S1, BMP2, BRDM2, BTR60, BTR70, D7, T62, T72, ZIL131, ZSU234). SOC trains at 17\ensuremath{^\circ} and tests at 15\ensuremath{^\circ}: 3,671 training and 3,203 test chips. We take the central 128\ensuremath{\times}128 pixels and use log magnitude.

\textbf{ATRNet-STAR.} The dataset contains 40 civilian vehicle models (13 cars, 8 buses and vans, 15 trucks and 4 special vehicles), imaged in X and Ku band, in four polarizations, at depression angles of 15\ensuremath{^\circ}, 30\ensuremath{^\circ}, 45\ensuremath{^\circ} and 60\ensuremath{^\circ}, in five scenes. Each chip is 128\ensuremath{\times}128 pixels and contains one vehicle with a bounding box; the median box is 45\ensuremath{\times}45 pixels. We use the official settings as distributed, without re-splitting (Table A). The dataset paper reports 194,324 images, but that total counts each image twice, once in ground-range and once in slant-range coordinates. We use the ground-range 8-bit amplitude images, 97,375 chips in the public release; the SOC-40 and EOC-scene settings divide these same chips in two different ways.

\begin{table*}[!t]
\centering
\footnotesize
\parbox{0.9\textwidth}{\centering \textbf{Table A.} ATRNet-STAR settings used in this paper.}\\[3pt]
\renewcommand{\arraystretch}{1.15}
\begin{tabular}{l>{\centering\arraybackslash}p{0.25\textwidth}>{\centering\arraybackslash}p{0.25\textwidth}>{\centering\arraybackslash}p{0.25\textwidth}}
\hline
\textbf{Setting} & \textbf{Train} & \textbf{Test} & \textbf{Used in} \\
\hline
SOC-40 & 68,091 & 29,284 & Section 4.2, Section 5, Section 6 \\
EOC-scene & 19,584 (sandstone 15,360, 40 classes; bare soil 4,224, 11 classes) & 77,791 (factory 31,843; urban 41,901; woodland 4,047, 11 classes) & Section 4.3 \\
\hline
\end{tabular}
\end{table*}

SOC-40 divides the chips 70/30 at random. EOC-scene divides them by scene: training uses the two simple scenes and testing the three cluttered ones. Most images were collected in the urban and factory scenes, so the EOC-scene test set is four times the size of its training set. SOC-40 by depression angle (train / test): 15\ensuremath{^\circ}, 15,831 / 6,689; 30\ensuremath{^\circ}, 17,379 / 7,493; 45\ensuremath{^\circ}, 17,389 / 7,558; 60\ensuremath{^\circ}, 17,492 / 7,544.

\subsection{Network and training}

Unless stated otherwise, every experiment uses ConvNeXt-Small with ImageNet weights \cite{liu2022}, a 128-pixel input, and the grey image copied to three channels. We train with AdamW \cite{loshchilov2019} (learning rate 1e-4, weight decay 0.05), a cosine schedule and batch size 128, with horizontal flips as augmentation; the masking experiments also shift images by up to 8 pixels. Each chip is contrast-stretched between its 1st and 99th percentiles. In the masking experiments the percentiles are computed on the visible pixels only, so masking does not change the intensity statistics of what remains. We train for 30 epochs on MSTAR, 10 on ATRNet-STAR SOC, 12 on EOC-scene and 15 in the angle experiments, and keep the epoch with the best accuracy on a 10\% validation split taken from the training set. Each test set is used once. For the masking experiments on ATRNet-STAR SOC, training is capped at 800 chips per class (28,800 training and 3,200 validation chips); the full official test set is used. Because of this cap, our full-chip SOC-40 accuracy (90.8\%) is below the 96.0\% the dataset authors report for ConvNeXt with all training data \cite{liu2026}. The comparisons in this paper are between inputs trained the same way, so the cap affects all of them equally.

\subsection{Reporting}

Every result is the mean and standard deviation over three seeds, and chance level is always shown. We also computed macro-F1; it agrees with accuracy within 1--3 points everywhere, so the tables show accuracy only. We never re-split chips at random: chips of the same vehicle at neighbouring azimuths are near-duplicates, and a random split would put them on both sides.

\section{Experiment 1: How Much Is Background?}

\subsection{Design}

We centre a square of side \emph{S} pixels on the vehicle. MSTAR vehicles are at the chip centre; for ATRNet-STAR we use the centre of the bounding box. We compare four inputs (Figure 1):

\begin{itemize}

\item
  \textbf{Full:} the whole chip.
\item
  \textbf{Background only:} the square is set to zero, so the vehicle is removed.
\item
  \textbf{Target only:} everything outside the square is set to zero.
\item
  \textbf{Mask position only:} the network sees only the binary map of which pixels are visible in the background-only input, with no image at all.
\end{itemize}

\emph{S} is the same for every chip, so the mask cannot reveal the vehicle's size. The mask-position input tests whether the vehicle's placement in the chip alone predicts its class. Training and test use the same masking.

\begin{figure}[!t]
\centering
\includegraphics[width=\columnwidth]{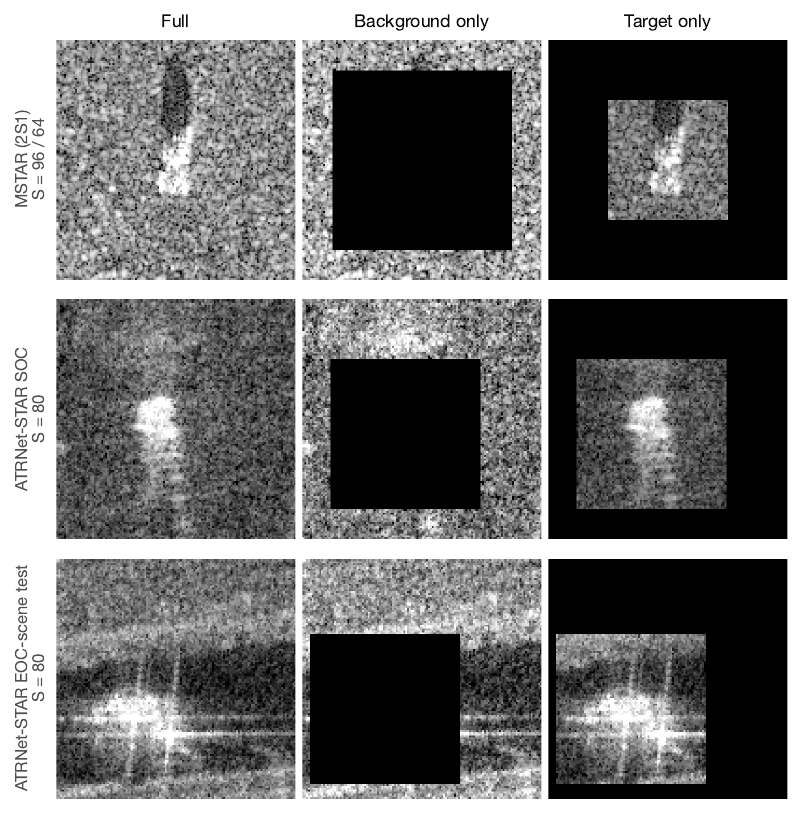}
\caption{Example inputs. Rows: an MSTAR test chip (2S1), an ATRNet-STAR SOC test chip and an ATRNet-STAR EOC-scene test chip from a cluttered scene. Columns: the full chip, the background-only input (the square around the vehicle set to zero) and the target-only input (everything outside the square set to zero). \emph{S} is the side of the square in pixels. On ATRNet-STAR the square follows the bounding box, so it is not always centred.}
\end{figure}

\subsection{Standard operating condition}

\begin{table*}[!t]
\centering
\footnotesize
\parbox{0.9\textwidth}{\centering \textbf{Table 1.} Test accuracy (\%), mean \ensuremath{\pm} std over 3 seeds, ConvNeXt-Small.}\\[3pt]
\renewcommand{\arraystretch}{1.15}
\begin{tabular}{lccccc}
\hline
\textbf{Dataset} & \textbf{Full} & \textbf{Background only} & \textbf{Target only} & \textbf{Mask position only} & \textbf{Chance} \\
\hline
MSTAR SOC & 99.1 \ensuremath{\pm} 0.1 & 74.9 \ensuremath{\pm} 0.6 (\emph{S} = 64); 73.3 \ensuremath{\pm} 1.3 (\emph{S} = 96) & 98.5 \ensuremath{\pm} 0.2 (\emph{S} = 64) & n/a (vehicles centred) & 10.0 \\
ATRNet-STAR SOC-40 & 90.8 \ensuremath{\pm} 0.1 & 63.3 \ensuremath{\pm} 0.3 (\emph{S} = 80) & 87.3 \ensuremath{\pm} 0.4 (\emph{S} = 80) & 5.2 \ensuremath{\pm} 0.4 & 2.5 \\
\hline
\end{tabular}
\end{table*}

On MSTAR the result matches earlier work. With the central 96\ensuremath{\times}96 pixels set to zero, only a 16-pixel frame of clutter remains, and the network still reaches 73.3\%, about seven times chance. Accuracy barely changes between \emph{S} = 64 and \emph{S} = 96 (74.9\% and 73.3\%), so the class information is spread through the clutter, not just in the area next to the vehicle. The same effect appears with a small CNN trained from scratch (Table S1), so it is not specific to a large pretrained network.

ATRNet-STAR was collected across several scenes, but within its SOC split the background is still highly informative: 63.3\% with the vehicle removed, 25 times chance. The vehicle alone is enough for a high score on both datasets (98.5\% and 87.3\%). It is just not needed for one. The mask-position input gives 5.2\% against 2.5\% chance, so vehicle placement leaks a little, but an order of magnitude less than the background.

\begin{table*}[!t]
\centering
\footnotesize
\parbox{0.9\textwidth}{\centering \textbf{Table S1.} Small CNN trained from scratch on MSTAR SOC, 3 seeds.}\\[3pt]
\renewcommand{\arraystretch}{1.15}
\begin{tabular}{p{3.2cm}>{\centering\arraybackslash}p{0.62\textwidth}}
\hline
\textbf{Input} & \textbf{Accuracy (\%)} \\
\hline
Full chip & 80.6 \ensuremath{\pm} 1.3 \\
Background only, \emph{S} = 48 / 64 / 80 / 96 / 112 & 75.6 \ensuremath{\pm} 1.5 / 68.4 \ensuremath{\pm} 0.9 / 64.0 \ensuremath{\pm} 0.6 / 61.8 \ensuremath{\pm} 2.8 / 39.9 \ensuremath{\pm} 1.9 \\
Target only, \emph{S} = 48 / 64 / 80 & 78.9 \ensuremath{\pm} 1.6 / 78.5 \ensuremath{\pm} 0.8 / 77.7 \ensuremath{\pm} 1.4 \\
\hline
\end{tabular}
\end{table*}

\subsection{Scene change}

\begin{table*}[!t]
\centering
\footnotesize
\parbox{0.9\textwidth}{\centering \textbf{Table 2.} ATRNet-STAR EOC-scene test accuracy (\%), 3 seeds, \emph{S} = 80. Training uses sandstone and bare soil; testing uses scenes not seen in training. The last column is validation accuracy on held-out chips from the training scenes.}\\[3pt]
\renewcommand{\arraystretch}{1.15}
\begin{tabular}{lccccc}
\hline
\textbf{Input} & \textbf{All test} & \textbf{Factory} & \textbf{Urban} & \textbf{Woodland*} & \textbf{Training scenes (val.)} \\
\hline
Full & 22.8 \ensuremath{\pm} 0.8 & 23.9 \ensuremath{\pm} 0.9 & 22.2 \ensuremath{\pm} 1.2 & 20.0 \ensuremath{\pm} 1.2 & 96.8 \ensuremath{\pm} 0.1 \\
Background only & 6.9 \ensuremath{\pm} 0.3 & 7.0 \ensuremath{\pm} 0.4 & 6.9 \ensuremath{\pm} 0.3 & 5.7 \ensuremath{\pm} 0.4 & 75.6 \ensuremath{\pm} 0.8 \\
\textbf{Target only} & \textbf{30.8 \ensuremath{\pm} 0.8} & \textbf{32.5 \ensuremath{\pm} 1.2} & \textbf{30.2 \ensuremath{\pm} 0.6} & \textbf{23.9 \ensuremath{\pm} 0.5} & 94.7 \ensuremath{\pm} 0.5 \\
Mask position only & 5.2 \ensuremath{\pm} 0.2 & 4.1 \ensuremath{\pm} 0.5 & 5.6 \ensuremath{\pm} 0.2 & 9.0 \ensuremath{\pm} 0.2 & 6.9 \ensuremath{\pm} 0.3 \\
\hline
\end{tabular}
\end{table*}

*Woodland contains only 11 of the 40 classes; chance over those classes is 9.1\%.

A change of scene is severe: 96.8\% on the training scenes and 22.8\% on new ones. Removing the background raises accuracy by 8 points, to 30.8\%, and the gain holds in all three test scenes. On new scenes, then, the background is not just useless; it misleads the network. Background-only accuracy confirms this. It is 75.6\% on the training scenes and 6.9\% on new ones, so the background signal in Table 1 belongs to the scenes it was learned in. The ATRNet-STAR authors suggested background interference as the cause of the EOC-scene drop; this measures it directly. A related observation on MSTAR, that training on the foreground alone can beat training on full chips, was reported in \cite{debiased2023}.

Our full-chip result is in line with the official EOC-scene baselines \cite{liu2026}, which range from 12.9\% to 33.6\% for magnitude-image networks, with 16.5\% for ConvNeXt and 19.5\% for SARATR-X. Our ConvNeXt reaches 22.8\%. Target-only input, at 30.8\%, is close to the best published magnitude-image result (33.6\%), without any change to the network.

\subsection{Does a SAR foundation model fix it?}

ConvNeXt was pretrained on photographs, so a network pretrained on SAR might attend to the vehicle instead. We repeat the controls with SARATR-X \cite{li2025}, a HiViT-Base encoder pretrained with masked image modelling on about 180,000 SAR images. We use the public checkpoint and fine-tune the whole network with a linear head. The only changes from Section 3.2 are that chips are upsampled from 128 to 224 pixels (the input size HiViT expects; masks are applied at 128 pixels first), the batch size is 64, and the first epoch uses one tenth of the learning rate as warm-up. Epochs, validation rule, seeds and inputs are otherwise the same.

\begin{table*}[!t]
\centering
\footnotesize
\parbox{0.9\textwidth}{\centering \textbf{Table 3.} Test accuracy (\%), mean \ensuremath{\pm} std over 3 seeds. The ConvNeXt column repeats Tables 1 and 2.}\\[3pt]
\renewcommand{\arraystretch}{1.15}
\begin{tabular}{lcccc}
\hline
\textbf{Dataset} & \textbf{Input} & \textbf{SARATR-X} & \textbf{ConvNeXt-Small} & \textbf{Difference} \\
\hline
MSTAR SOC & Full & 99.1 \ensuremath{\pm} 0.1 & 99.1 \ensuremath{\pm} 0.1 & 0.0 \\
MSTAR SOC & Background only (\emph{S} = 96) & \textbf{79.2 \ensuremath{\pm} 0.7} & 73.3 \ensuremath{\pm} 1.3 & \textbf{+5.9} \\
ATRNet-STAR SOC-40 & Full & 91.3 \ensuremath{\pm} 0.3 & 90.8 \ensuremath{\pm} 0.1 & +0.5 \\
ATRNet-STAR SOC-40 & Background only (\emph{S} = 80) & \textbf{67.5 \ensuremath{\pm} 0.4} & 63.3 \ensuremath{\pm} 0.3 & \textbf{+4.2} \\
EOC-scene & Full & \textbf{16.8 \ensuremath{\pm} 0.9} & 22.8 \ensuremath{\pm} 0.8 & \textbf{\ensuremath{-}6.0} \\
EOC-scene & Target only (\emph{S} = 80) & 26.1 \ensuremath{\pm} 0.6 & 30.8 \ensuremath{\pm} 0.8 & \ensuremath{-}4.7 \\
EOC-scene & Background only (\emph{S} = 80) & 6.0 \ensuremath{\pm} 0.1 & 6.9 \ensuremath{\pm} 0.3 & \ensuremath{-}0.9 \\
\hline
\end{tabular}
\end{table*}

SAR pretraining does not remove the background shortcut. It makes it stronger. With the vehicle removed, SARATR-X is 4 to 6 points more accurate than ConvNeXt on both benchmarks, in every seed, while its full-chip accuracy is the same. The extra reliance on background is invisible in the standard benchmark number. Under the scene change, SARATR-X reaches 16.8\% on full chips, 6 points below our ConvNeXt, and target-only input again beats full chips (26.1\% against 16.8\%). We do not read much into the ranking of the two networks here. The official benchmark reports the opposite order (19.5\% for SARATR-X and 16.5\% for ConvNeXt \cite{liu2026}), so under a scene change the ranking depends on the training recipe. What holds in both is that SAR pretraining does not solve the scene change. The background-only result above was not part of the official benchmark, and it holds in every seed and on both datasets.

One possible explanation is that masked image modelling on SAR teaches the encoder to model terrain and clutter texture well. That is exactly the signal that separates classes within one collection and fails across scenes. We have not tested this. The comparison also has limits: one foundation model, one fine-tuning recipe, chips upsampled from 128 pixels while SARATR-X was pretrained at 224, and no hyper-parameter search for either network. A different recipe might narrow the background-only gap. Given that it holds in every seed and on both datasets, we do not expect it to reverse.

\section{Experiment 2: Depression-Angle Gap at a Fixed Data Budget}

\subsection{Design}

We choose a set \emph{A} of training angles from \{15\ensuremath{^\circ}, 30\ensuremath{^\circ}, 45\ensuremath{^\circ}, 60\ensuremath{^\circ}\} and run all 14 non-empty proper subsets. Each network is trained on ATRNet-STAR SOC-40 training chips at the angles in \emph{A} and tested on the official SOC-40 test chips at every angle not in \emph{A}.

The amount of training data is fixed: every condition gets 230 chips per class, divided equally over the angles in \emph{A} (8,280 training and 920 validation chips in total). Adding angles therefore never adds data, and any difference between conditions comes from which angles the data were taken at. The figure of 230 is set by the smallest class-and-angle group, which has 236 chips at 15\ensuremath{^\circ}. Each test angle is labelled by its gap to the nearest training angle, and as interpolation (between two training angles) or extrapolation. Standard deviations across the three seeds average 0.7 points and never exceed 1.5.

\subsection{Results}

\begin{table*}[!t]
\centering
\footnotesize
\parbox{0.9\textwidth}{\centering \textbf{Table 4.} Test accuracy (\%) at each held-out angle, mean \ensuremath{\pm} std over 3 seeds; chance 2.5\%. \emph{i} marks interpolation; all other cells are extrapolation. The last column is validation accuracy at the training angles.}\\[3pt]
\renewcommand{\arraystretch}{1.15}
\begin{tabular}{lccccc}
\hline
\textbf{Train angles} & \textbf{Test 15\ensuremath{^\circ}} & \textbf{Test 30\ensuremath{^\circ}} & \textbf{Test 45\ensuremath{^\circ}} & \textbf{Test 60\ensuremath{^\circ}} & \textbf{Val. at train angles} \\
\hline
15 &  & 54.6 \ensuremath{\pm} 0.5 & 41.0 \ensuremath{\pm} 0.4 & 22.4 \ensuremath{\pm} 0.3 & 88.0 \\
30 & 58.5 \ensuremath{\pm} 0.2 &  & 71.0 \ensuremath{\pm} 0.4 & 30.0 \ensuremath{\pm} 0.6 & 87.8 \\
45 & 47.4 \ensuremath{\pm} 0.6 & 71.9 \ensuremath{\pm} 0.4 &  & 41.6 \ensuremath{\pm} 0.8 & 89.2 \\
60 & 29.7 \ensuremath{\pm} 0.7 & 38.7 \ensuremath{\pm} 1.1 & 49.8 \ensuremath{\pm} 1.5 &  & 82.8 \\
15+30 &  &  & 63.9 \ensuremath{\pm} 1.3 & 29.7 \ensuremath{\pm} 0.7 & 82.2 \\
15+45 &  & 72.5 \ensuremath{\pm} 0.5 \emph{i} &  & 38.6 \ensuremath{\pm} 0.6 & 81.1 \\
15+60 &  & 57.9 \ensuremath{\pm} 0.4 \emph{i} & 56.9 \ensuremath{\pm} 0.6 \emph{i} &  & 72.6 \\
30+45 & 55.7 \ensuremath{\pm} 1.5 &  &  & 39.0 \ensuremath{\pm} 1.4 & 85.1 \\
30+60 & 54.9 \ensuremath{\pm} 0.7 &  & 71.3 \ensuremath{\pm} 1.0 \emph{i} &  & 76.5 \\
45+60 & 43.8 \ensuremath{\pm} 0.6 & 64.3 \ensuremath{\pm} 0.9 &  &  & 76.4 \\
15+30+45 &  &  &  & 36.7 \ensuremath{\pm} 1.0 & 77.4 \\
15+30+60 &  &  & 66.6 \ensuremath{\pm} 0.6 \emph{i} &  & 71.0 \\
15+45+60 &  & 68.1 \ensuremath{\pm} 0.8 \emph{i} &  &  & 72.8 \\
30+45+60 & 53.8 \ensuremath{\pm} 0.7 &  &  &  & 76.3 \\
\hline
\end{tabular}
\end{table*}
\begin{table*}[!t]
\centering
\footnotesize
\parbox{0.9\textwidth}{\centering \textbf{Table 5.} Table 4 summarized by the gap to the nearest training angle. Each condition is one cell of Table 4.}\\[3pt]
\renewcommand{\arraystretch}{1.15}
\begin{tabular}{lcccc}
\hline
\textbf{Gap} & \textbf{Training angles} & \textbf{Conditions} & \textbf{Mean accuracy} & \textbf{Range} \\
\hline
15\ensuremath{^\circ} & 1 & 6 & 57.9 & 41.6--71.9 \\
15\ensuremath{^\circ} & 2 & 10 & 57.5 & 38.6--72.5 \\
15\ensuremath{^\circ} & 3 & 4 & 56.3 & 36.7--68.1 \\
30\ensuremath{^\circ} & 1 & 4 & 39.3 & 30.0--47.4 \\
30\ensuremath{^\circ} & 2 & 2 & 36.8 & 29.7--43.8 \\
45\ensuremath{^\circ} & 1 & 2 & 26.0 & 22.4--29.7 \\
\hline
\end{tabular}
\end{table*}
\begin{figure}[!t]
\centering
\includegraphics[width=0.95\columnwidth]{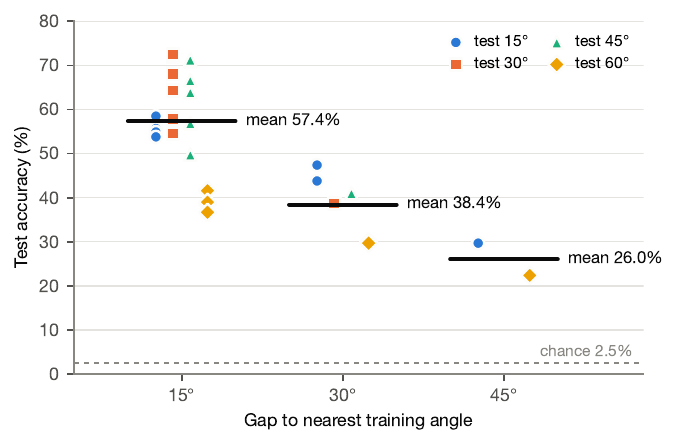}
\caption{Accuracy against the gap to the nearest training angle. Each point is one cell of Table 4 (3-seed mean), and the shape and colour give the test angle. Points are offset sideways slightly so they do not overlap. Black bars are the means over all conditions at each gap.}
\end{figure}

\begin{figure}[!t]
\centering
\includegraphics[width=0.95\columnwidth]{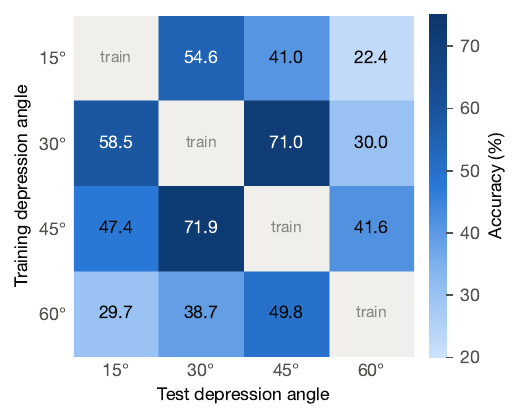}
\caption{Accuracy when training at a single depression angle (rows) and testing at another (columns): the first four rows of Table 4. Neighbouring middle angles (30\ensuremath{^\circ} and 45\ensuremath{^\circ}) transfer best; 60\ensuremath{^\circ} is hardest to reach.}
\end{figure}

\subsection{Findings}

The gap to the nearest training angle largely sets the accuracy. The mean is 57\% at a 15\ensuremath{^\circ} gap, 39\% at 30\ensuremath{^\circ} and 26\% at 45\ensuremath{^\circ}, a loss of 15 to 20 points for each 15\ensuremath{^\circ} step. At the training angles themselves, validation accuracy is 83--89\%.

With the amount of data fixed, more training angles do not help. At a 15\ensuremath{^\circ} gap the mean is 57.9\% with one training angle, 57.5\% with two and 56.3\% with three. Bracketing the test angle adds little: at 45\ensuremath{^\circ}, training at 30\ensuremath{^\circ} alone gives 71.0\%, and training at 30\ensuremath{^\circ} and 60\ensuremath{^\circ} gives 71.3\%. Adding a distant angle can even hurt. At 45\ensuremath{^\circ}, training at 30\ensuremath{^\circ} alone gives 71.0\%, while 15\ensuremath{^\circ}+30\ensuremath{^\circ} gives 63.9\% and 15\ensuremath{^\circ}+30\ensuremath{^\circ}+60\ensuremath{^\circ} gives 66.6\%. Splitting the budget takes chips away from the angle that matters most.

The middle angles transfer best. Between 30\ensuremath{^\circ} and 45\ensuremath{^\circ} accuracy is 71--72\% in both directions. The steep end is hardest: the best result at 60\ensuremath{^\circ} is 41.6\%, from training at 45\ensuremath{^\circ} alone. Transfer is also not symmetric. Training at 60\ensuremath{^\circ} and testing at 45\ensuremath{^\circ} gives 49.8\%, while the reverse gives 41.6\%; 30\ensuremath{^\circ} to 15\ensuremath{^\circ} gives 58.5\%, and 15\ensuremath{^\circ} to 30\ensuremath{^\circ} gives 54.6\%. A plausible physical reason is that shadow length, layover and the dominant scatterers all change with depression angle, and change fastest at the steep end. We have not tested this.

Validation accuracy at the training angles also falls as angles are mixed: 88\% with one angle and 71--85\% with two or three. With a fixed budget, each angle gets fewer chips, and the task within each angle becomes harder.

These conclusions hold at this data budget. With much more data at every angle, adding angles might help; we did not test that.

\section{Experiment 3: The Partial-Library Effect}

\subsection{Design}

In practice a reference library holds most vehicles at most angles, with some gaps. We simulate this. Networks train at two angles, \emph{A} and \emph{B}, and are tested at \emph{B}. The 40 classes are split into four fixed folds of 10. Classes in the held-out fold get all 230 chips at \emph{A} only. The other 30 classes get 115 chips at \emph{A} and 115 at \emph{B}, the same total per class. Running all four folds makes every class ``missing'' exactly once, so the pooled accuracy on missing classes can be compared directly with the matching cell of Table 4, where \emph{every} class was missing at \emph{B}.

We also count how often a missing class's test chips are assigned to one of the 30 library classes. If the network guessed uniformly over all 40 classes, 75\% (30/40) of its wrong guesses would land on a library class.

\subsection{Results}

\begin{table*}[!t]
\centering
\footnotesize
\parbox{0.9\textwidth}{\centering \textbf{Table 6.} Accuracy (\%) on the 10 classes missing at the test angle, pooled over 4 folds, mean \ensuremath{\pm} std over 3 seeds.}\\[3pt]
\renewcommand{\arraystretch}{1.15}
\begin{tabular}{lccccc}
\hline
\textbf{Train \ensuremath{\rightarrow} test} & \textbf{Gap} & \textbf{No class at \emph{B} (Table 4)} & \textbf{Missing classes, others at \emph{B}} & \textbf{Library classes at \emph{B}} & \textbf{Missing chips \ensuremath{\rightarrow} library class} \\
\hline
30 \ensuremath{\rightarrow} 45 & 15\ensuremath{^\circ} & 71.0 & 68.6 \ensuremath{\pm} 0.6 & 85.6 \ensuremath{\pm} 0.4 & 26.4\% \\
15 \ensuremath{\rightarrow} 45 & 30\ensuremath{^\circ} & 41.0 & \textbf{28.7 \ensuremath{\pm} 1.2} & 81.1 \ensuremath{\pm} 0.7 & 68.7\% \\
15 \ensuremath{\rightarrow} 60 & 45\ensuremath{^\circ} & 22.4 & \textbf{9.0 \ensuremath{\pm} 0.4} & 68.6 \ensuremath{\pm} 0.5 & 89.5\% \\
\hline
\end{tabular}
\end{table*}

Across the 12 fold-and-seed runs, missing-class accuracy ranges over 66.7--70.7\%, 18.5--40.5\% and 2.7--13.1\% for the three rows. Variation between seeds is small; most of the spread comes from which classes are missing.

\begin{figure}[!t]
\centering
\includegraphics[width=0.95\columnwidth]{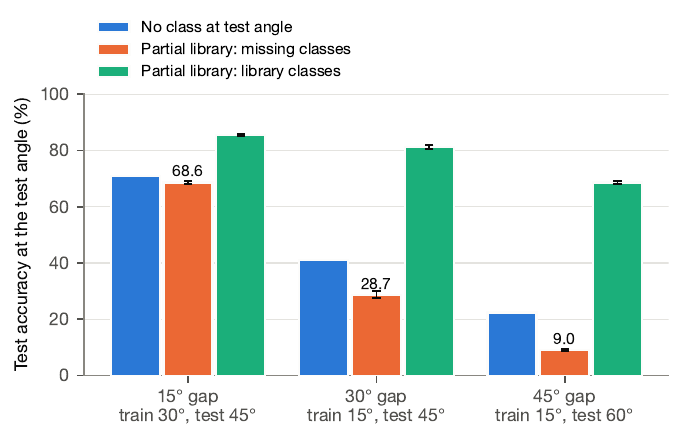}
\caption{The partial-library effect (Table 6). Blue: accuracy when no class has training chips at the test angle. Orange: accuracy on the classes missing from an otherwise complete library at that angle. Green: accuracy on the library classes. Error bars are the standard deviation over 3 seeds.}
\end{figure}

\subsection{Findings}

A partial library hurts the classes it lacks. At a 30\ensuremath{^\circ} gap, missing-class accuracy falls from 41.0\% to 28.7\%; at 45\ensuremath{^\circ} it falls from 22.4\% to 9.0\%, and to 2.7\% in one fold. At a 15\ensuremath{^\circ} gap the effect is small (71.0\% to 68.6\%). The drops hold in every seed.

The missing classes are absorbed by the library. At a 45\ensuremath{^\circ} gap, 89.5\% of missing-class chips are assigned to a library class, more than the 75\% expected from uniform guessing. The network seems to learn that ``looks like angle \emph{B}'' means ``one of the classes seen at \emph{B}'', so the appearance of the angle itself becomes a class cue.

Meanwhile the library classes do well at \emph{B} (69--86\%), so a user would see a network that seems competent at that angle. A gap in the library does not produce ``I don't know''. It produces a wrong answer that names a known vehicle, and the larger the angle gap, the more often this happens.

\section{Discussion}

\subsection{What a high benchmark score shows}

A score near 99\% on MSTAR SOC, or above 90\% on ATRNet-STAR SOC, shows that a network separates the classes under the conditions of that collection. It does not show how much of that comes from the vehicle. In both benchmarks a large part of the score can be reached with the vehicle removed. Under a real change of scene, the same background cue turns against the network, and a SAR-pretrained network relies on it more, not less. None of this is visible in the headline number.

\subsection{What we recommend reporting}

The controls in this paper are cheap: two extra training runs per setting. We suggest that SAR ATR papers report, next to every SOC or EOC accuracy:

\begin{itemize}

\item
  background-only and target-only accuracy, with the mask size;
\item
  a mask-position control, when vehicles are not centred;
\item
  for any angle result, the gap to the nearest training angle and whether the amount of training data was held fixed.
\end{itemize}

Our code produces all of these.

\subsection{Operational meaning}

Fielded systems usually know the collection geometry and compare against a reference library. Our angle results suggest how dense that library must be: accuracy falls by 15 to 20 points for every 15\ensuremath{^\circ} of depression-angle gap, and faster at the steep end. The partial-library result adds a warning. A gap in the library does not show up as low confidence or as an ``unknown'' output. It shows up as a confident identification of a different, known vehicle.

Because the collection angle is always known, this suggests a simple rule: compute the gap between the image's angle and the nearest library entry for each candidate class, and defer to an analyst when that gap is large. This does not need a better network, only the metadata that is already available. Simulation tools such as CAD-based SAR modelling could help fill library gaps, but simulated and measured SAR still differ, and that gap would need its own controls.

A natural next step is to raise the vehicle's weight relative to the clutter during training, for example by attenuating the background in steps, pasting vehicles onto other backgrounds, or penalizing attention outside the vehicle box, and to measure which of these best survives a change of scene. We leave this to follow-up work.

The same kind of shortcut appears in our parallel work on ships in Sentinel-1 imagery matched to AIS. There, location and ship size explain much of what image models learn. We will report that separately.

\subsection{Limitations}

\begin{itemize}

\item
  We use magnitude chips only, without phase or polarimetric information.
\item
  Zero-filling is itself a change in the input. We apply the same masking in training and test to limit its effect.
\item
  The main results use one architecture (ConvNeXt-Small) and one SAR foundation model (SARATR-X), each with one training recipe. A small CNN on MSTAR (Table S1) shows the same background effect.
\item
  ATRNet-STAR SOC training is capped at 800 chips per class in Section 4, and the angle experiments use 230 chips per class.
\item
  Depression angles are available only in 15\ensuremath{^\circ} steps, so we cannot measure accuracy below a 15\ensuremath{^\circ} gap.
\item
  The collection details that might explain the background correlation in MSTAR are not public.
\item
  The partial-library experiment uses one fold design (4 \ensuremath{\times} 10 classes) and three angle pairs.
\end{itemize}

\section{Conclusion}

We asked how much of SAR vehicle recognition is the vehicle. On MSTAR we confirm, with a modern network, what has been known since 2005: much of the score survives with the vehicle removed. The newer ATRNet-STAR benchmark has the same shortcut, it reverses under a change of scene, and a SAR foundation model leans on it more than a network pretrained on photographs. Depression-angle gaps cost 15 to 20 points per 15\ensuremath{^\circ}, more training angles do not compensate at a fixed budget, and a partly filled library turns missing vehicles into confident misidentifications. These effects are easy to measure, and we suggest that they be reported alongside any accuracy figure.

\section*{Code and Data}
\addcontentsline{toc}{section}{Code and Data}

Code: https://github.com/mousashaya/sar-atr-study

\begin{itemize}

\item
  Background controls: \texttt{scripts/train\_mask.py}, \texttt{scripts/build\_atr\_cache.py} (Section 4)
\item
  Angle gap: \texttt{scripts/train\_angle\_all.py}, \texttt{scripts/train\_angle\_heldout.py} (Section 5)
\item
  Partial library: \texttt{scripts/train\_angle\_heldout\_class.py} (Section 6)
\end{itemize}

MSTAR is available from AFRL SDMS. ATRNet-STAR is available from its authors.

\section*{Acknowledgment}
Code implementation and drafting of the text were assisted by Claude (Anthropic). The experimental design, the interpretation and all claims are the author's responsibility.


\begin{thebibliography}{99}
\bibitem{geirhos2020} R. Geirhos \emph{et al.}, ``Shortcut learning in deep neural networks,'' \emph{Nature Machine Intelligence}, vol.~2, pp.~665--673, 2020.
\bibitem{schumacher2004} R. Schumacher and K. Rosenbach, ``ATR of battlefield targets by SAR: classification results using the public MSTAR dataset compared with a dataset by QinetiQ UK,'' \emph{RTO SET Symposium on Target Identification and Recognition Using RF Systems}, 2004.
\bibitem{schumacher2005} R. Schumacher and J. Schiller, ``Non-cooperative target identification of battlefield targets -- classification results based on SAR images,'' \emph{Proc. IEEE International Radar Conference}, 2005, pp.~167--172.
\bibitem{heiligers2018} M. Heiligers and A. Huizing, ``On the importance of visual explanation and segmentation for SAR ATR using deep learning,'' \emph{IEEE Radar Conference (RadarConf18)}, 2018, pp.~394--399.
\bibitem{zhou2018} F. Zhou, L. Wang, X. Bai, Y. Hui, ``SAR ATR of ground vehicles based on LM-BN-CNN,'' \emph{IEEE Trans. Geosci. Remote Sens.}, vol.~56, no. 12, pp.~7282--7293, 2018.
\bibitem{belloni2021} C. Belloni, A. Balleri, N. Aouf, J.-M. Le Caillec, T. Merlet, ``Explainability of deep SAR ATR through feature analysis,'' \emph{IEEE Trans. Aerosp. Electron. Syst.}, vol.~57, no. 1, pp.~659--673, 2021.
\bibitem{li2023} W. Li, W. Yang, L. Liu, W. Zhang, Y. Liu, ``Discovering and explaining the noncausality of deep learning in SAR ATR,'' \emph{IEEE Geosci. Remote Sens. Lett.}, vol.~20, 2023, doi:10.1109/LGRS.2023.3266493.
\bibitem{liu2026} Y. Liu, W. Li, L. Liu, J. Zhou, B. Peng \emph{et al.}, ``ATRNet-STAR: A large dataset and benchmark towards remote sensing object recognition in the wild,'' \emph{IEEE Trans. Pattern Anal. Mach. Intell.}, 2026, doi:10.1109/TPAMI.2026.3658649.
\bibitem{li2025} W. Li, W. Yang, Y. Hou, L. Liu, Y. Liu, X. Li, ``SARATR-X: Toward building a foundation model for SAR target recognition,'' \emph{IEEE Trans. Image Process.}, vol.~34, pp.~869--884, 2025, doi:10.1109/TIP.2025.3531988.
\bibitem{ross1998} T. D. Ross, S. W. Worrell, V. J. Velten, J. C. Mossing, M. L. Bryant, ``Standard SAR ATR evaluation experiments using the MSTAR public release data set,'' \emph{Proc. SPIE}, vol.~3370, pp.~566--573, 1998, doi:10.1117/12.321859.
\bibitem{geng2023} Z. Geng, Y. Xu, B.-N. Wang, X. Yu, D.-Y. Zhu, G. Zhang, ``Target recognition in SAR images by deep learning with training data augmentation,'' \emph{Sensors}, vol.~23, no. 2, art. 941, 2023.
\bibitem{peng2023} B. Peng, J. Xie, B. Peng, L. Liu, ``Learning invariant representation via contrastive feature alignment for clutter robust SAR ATR,'' \emph{IEEE Geosci. Remote Sens. Lett.}, vol.~20, 2023 (arXiv:2304.01747).
\bibitem{debiased2023} H. Dong, F. Han, L. Si, W. Qiang, L. Zhang, ``Background debiased SAR target recognition via causal interventional regularizer,'' arXiv:2308.15724, 2023.
\bibitem{li2024} W. Li \emph{et al.}, ``Predicting gradient is better: Exploring self-supervised learning for SAR ATR with a joint-embedding predictive architecture,'' \emph{ISPRS J. Photogramm. Remote Sens.}, vol.~218, pp.~326--338, 2024, doi:10.1016/j.isprsjprs.2024.09.013.
\bibitem{jiang2025} C. Jiang, C. Wang, F. Wu, P. Ma, L. Zou, T. Li, J. Ning, Y. Tang, ``SARCLIP: A multimodal foundation framework for SAR imagery via contrastive language-image pre-training,'' \emph{ISPRS J. Photogramm. Remote Sens.}, vol.~231, pp.~17--34, 2026, doi:10.1016/j.isprsjprs.2025.10.017.
\bibitem{liu2022} Z. Liu, H. Mao, C.-Y. Wu, C. Feichtenhofer, T. Darrell, S. Xie, ``A ConvNet for the 2020s,'' \emph{Proc. IEEE/CVF CVPR}, 2022.
\bibitem{loshchilov2019} I. Loshchilov and F. Hutter, ``Decoupled weight decay regularization,'' \emph{Proc. ICLR}, 2019.
\end{thebibliography}
\end{document}